**Title:** Patient-Centered Summarization Framework for AI Clinical Summarization: A Mixed-Methods Design

**Maria Lizarazo Jimenez[1*], Ana Gabriela Claros[1*], Kieran Green[2], David Toro-Tobon[1,3], Felipe Larios[1], Sheena Asthana[4], Camila Wenczenovicz[1], Kerly Guevara Maldonado[1], Luis Vilatuna-Andrango[1], Cristina Proano-Velez[1], Satya Sai Sri Bandi[1], Shubhangi Bagewadi[1], Megan E. Branda[1,5], Misk Al Zahidy[1], Saturnino Luz[6], Mirella Lapata[7], Juan P. Brito[1,3], Oscar J. Ponce-Ponte[1,2]**

1. Care and AI Laboratory, Knowledge and Evaluation Research Unit, Division of Endocrinology, Diabetes, Metabolism and Nutrition, Department of Medicine, Mayo Clinic, Rochester, Minnesota, USA
2. University Hospitals Plymouth NHS Trust, Derriford Hospital, Plymouth, UK
3. Division of Endocrinology, Diabetes, Metabolism, and Nutrition, Mayo Clinic, Rochester, Minnesota, USA
4. Centre for Health Technology, University of Plymouth, UK
5. Division of Clinical Trials and Biostatistics, Department of Quantitative Health Sciences, Mayo Clinic, Rochester, Minnesota, USA.
6. Usher Institute, University of Edinburgh
7. School of Informatics, University of Edinburgh

*These authors contributed equally.

**Corresponding author:**

Oscar J. Ponce-Ponte

ponceponte.oscar@mayo.edu

Care and AI Laboratory, Knowledge and Evaluation Research Unit, Division of Endocrinology, Diabetes, Metabolism and Nutrition, Department of Medicine, Mayo Clinic, Rochester, Minnesota, USA

Mayo Clinic, 200 First Street SW, Rochester, MN 55902.

Word count: 3,751; Figures: 2; Tables: 3

Large Language Models (LLMs) are increasingly demonstrating the potential to reach human-level performance in generating clinical summaries from patient-clinician conversations. However, these summaries often focus on patients' biology rather than their preferences, values, wishes, and concerns. To achieve patient-centered care, we propose a new standard for Artificial Intelligence (AI) clinical summarization tasks: Patient-Centered Summaries (PCS). Our objective was to develop a framework to generate PCS that capture patient values and ensure clinical utility, and to assess whether current open-source LLMs can achieve human-level performance in this task. We used a mixed-methods process. Two Patient and Public Involvement groups (10 patients and 8 clinicians) in the United Kingdom participated in semi-structured interviews exploring what personal and contextual information should be included in clinical summaries and how it should be structured for clinical use. Findings informed annotation guidelines used by eight clinicians to create gold-standard PCS from 88 atrial fibrillation consultations. Sixteen consultations were used to refine a prompt aligned with the guidelines. Five open-source LLMs (Llama-3.2-3B, Llama-3.1-8B, Mistral-8B, Gemma-3-4B, and Qwen3-8B) generated summaries for 72 consultations using zero-shot and few-shot prompting, evaluated with ROUGE-L, BERTScore, and qualitative metrics. Patients emphasized lifestyle routines, social support, recent stressors, and care values. Clinicians sought concise functional, psychosocial, and emotional context. The best zero-shot performance was achieved by Mistral-8B (ROUGE-L 0.189) and Llama-3.1-8B (BERTScore 0.673); the best few-shot by Llama-3.1-8B (ROUGE-L 0.206, BERTScore 0.683). Completeness and fluency were similar between experts and models, while correctness and patient-centeredness favored human PCS.

## INTRODUCTION

Clinical summarization is an essential component of modern healthcare, as it allows clinicians to efficiently aggregate, organize, and synthesize complex patient data into concise, actionable insights that support clinical reasoning, decision-making, and care coordination[1]. High-quality summaries help reduce information overload, promote safer care transitions, and improve communication among providers and with patients[1-3]. For example, high-quality discharge summaries have been associated with fewer medication errors, lower readmission rates, mitigating diagnostic errors, facilitating effective communication between care teams, and increasing clinician satisfaction[2,4]. Yet preparing these summaries is time-consuming and contributes to documentation burden, which has motivated the development of artificial intelligence (AI) systems to automate the task. These systems use natural language processing to generate concise summaries from electronic health records (EHRs) and patient-clinician encounters [5,6]. Recent work in clinical dialogue summarization has employed multi-stage pipelines and fine-tuned large language models (LLMs) to improve the accuracy and coherence of patient-clinician summaries, reaching almost human-level performance[6].

However, current AI models are trained on clinical notes that have increasingly evolved to capture biological complexity as well as provide enough documentation to justify billing and reimbursement[7]. This dual purpose has shaped current summaries and the datasets used to guide current AI scribes, making them biologically centered and prioritizing pathophysiological details over patient preferences, their context, and what truly matters to them[8]. As a result, these models often produce clinical summaries that focus on medical issues and the clinician's perspective, without fully incorporating patients' values, preferences, and concerns. Therefore, we posit that patient-centered summarization is a new AI task that captures patients' values and preferences while preserving clinical utility. This a key issue for enabling patient-centered pathways particularly for older patients with multimorbidity, whose experience of clinical encounters may be less shaped by the need for condition-specific information than a holistic approach that understands their needs to manage complex health problems to maximize quality of life[9,10].

To ensure that patient-centered summaries (PCS) truly capture what matters to patients while remaining clinically useful, their definition must be dynamic and context-dependent. What constitutes a "patient-centered" summary may vary across settings, medical specialties, cultural contexts, geographic regions, and individual backgrounds. Therefore, PCS must continuously adapt to both patients' and clinicians' perspectives within these varying contexts. Achieving this requires ongoing collection and synthesis of feedback from both groups to define an evolving gold-standard PCS, which can then guide local benchmarking or fine-tuning of AI systems.

Because this process is often time-consuming and resource-intensive, we propose a methodology to accelerate and streamline it through patient and public involvement and engagement (PPIE) groups. In this study, we demonstrate the feasibility of this approach, including the benchmarking of open-source large language models (LLMs). Specifically, we (1) introduce a novel mixed-methods framework for developing a context-aware, patient-centered, gold-standard for clinical summarization, and (2) benchmark current open-source LLMs against gold-standard PCS to establish a baseline for future improvement in this area.

## METHODS

The framework to define, generate, and evaluate PCS followed a four-step a mixed-methods design. First, we captured patients' and clinicians' perspectives into PCS by leveraging the Patient and Public Involvement and Engagement (PPIE) methodology. Two PPIE groups were convened to define the core components of a PCS through semi-structured interviews. Second, these findings were translated into an annotation guideline, which eight clinicians' annotators used to create gold-standard PCS for 88 transcribed patient-clinician consultations. Third, a prompt was iteratively developed using 16 consultations and then used to generate PCS from five general-purpose, open-source LLMs (Llama-3.2-3B, Llama-3.1-8B, Mistral-8B, Gemma-3-4B, and Qwen3-8B) with zero-shot and few-shot prompting techniques. Notably, none of these models were pretrained on clinical summaries from EHRs. Finally, the model-generated summaries from the remaining 72 consultations were evaluated against the gold-standard PCS using a quantitative analysis (ROUGE-L and BERTScore) and a qualitative human assessment across five domains, including correctness and patient-centeredness. **Figure 1** provides an overview of the study workflow.

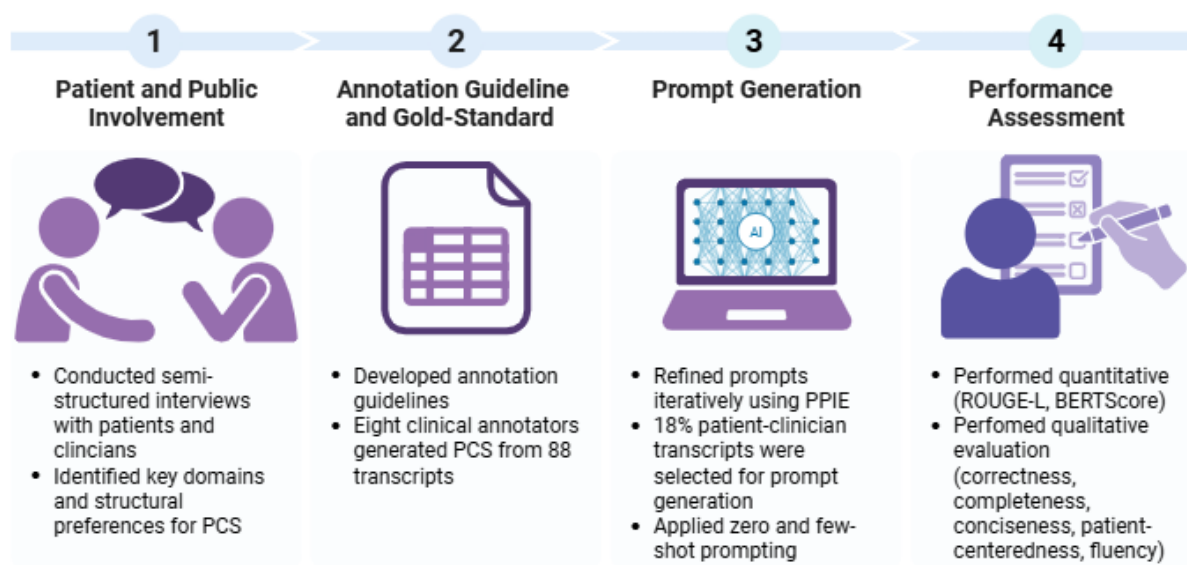

**Figure 1. Study Methods. The study followed four phases: (1) Patient and Public Involvement and Engagement (PPIE) to identify key domains for patient-centered summaries (PCS); (2) development of annotation guidelines and a consensus-based gold-standard PCS; (3) prompt generation refined with PPIE and tested in zero- and few-shot settings; and (4) performance assessment using quantitative metrics (ROUGE-L, BERTScore) and qualitative human evaluation.**

First step: Patient and Public Involvement

Two PPIE groups were recruited from Devon, a southwest coastal region in the United Kingdom. Eligible participants included adult patients (aged 18 or older) with capacity to provide informed consent and engage in PPIE discussions, as well as clinicians who regularly conducted face-to-face clinical appointments. Patients were excluded if they lacked decision-making capacity or could not meaningfully participate. Clinicians were excluded if they did not routinely provide direct patient care, meaning those with continuous contact with patients in the outpatient clinic.

Semi-structured interviews for the PPIEs were led by K.G, who developed and iteratively improved and simplified the interview questions with input from experts in patient-centeredness. These questions explored the importance of capturing 1) Personal and Lifestyle information, 2) Support systems and Beliefs, 3) Health Goals and Challenges, 4) Emotional and Mental Health, 5) Sources of Meaning, and for clinicians, an additional question about how they preferred to receive and prioritize the information discussed from an appointment. These domains were derived and synthesized from integrative models of patient-centered care and from expert-validated frameworks that prioritize dimensions such as the unique person, patient involvement, provision of information, clinician–patient communication, and empowerment[11,12]. These questions were implemented so that patients discussed what personal and contextual information should be captured in clinical summaries, while clinicians reflected on how such information could be structured for clinical use.

Interview notes were handwritten during or shortly after sessions and thematically coded using the five patient-centered domains, along with an additional category for structural preferences. Data were stratified by participant age group: young (18–30 years), middle-aged (31–60 years), and older (61-91) and role (patient or clinicians) and analyzed through iterative thematic analysis to identify common and divergent priorities. Relevant quotes were categorized and introduced into each subheading.

Second step: Annotation guideline and gold-standard PCS

The results from the first step were used to develop the annotation guideline, which incorporated input from both the clinicians' PPIE to guide the desired format and the patients' PPIE to provide illustrative quotes and categories within each key area, as shown in **Supplementary Material 1.**

A team of eight clinician annotators applied the guideline after a calibration phase to ensure shared understanding. In this phase, pairs worked on the same transcript, identifying background topics, issues, the plan, and patient-centered elements, and generating summaries until their outputs showed similar structure and complete inclusion of PCS elements. Disagreements were resolved through discussion. After calibration, annotators worked independently; some transcripts were reviewed by more than one annotator, but no formal statistic of inter-annotator agreement was calculated, as differences were largely stylistic. Instead, consistency was reinforced in team meetings, where summaries were compared to confirm shared standards and inclusion of PCS elements.

Annotators generated 88 PCS from non-scripted patient-clinician consultations obtained from a shared decision-making clinical trial in atrial fibrillation[13]. A total of 922 encounters were originally recorded. Scripted encounters from the intervention arm were excluded because clinicians followed the decision aid step by step, limiting exploration of other aspects of patients' lives. Encounters were also excluded when essential components of the consultation structure such as a clear discussion and agreement on a management plan were missing. After applying these criteria, the team purposively selected 88 eligible encounters to ensure diversity in patient and clinician characteristics as well as in communication styles. Original clinical notes were not used as a comparator, as they are clinician-oriented and focus primarily on biomedical aspects. In contrast, transcriptions captured the full dialogue, including patients' perspectives and contextual details, providing a comprehensive and ethically accessible source for developing patient-centered summaries.

Third step: prompt generation

Out of 88 transcribed patient-clinician conversations, we selected 16 (18%) for prompt generation. The prompt **Supplementary Material 2** was iteratively improved based on the annotation guideline. The instruction on how to generate each section of the clinical summary was iteratively improved based on qualitative similarity between the AI-generated summary and the gold-standard PCS and quantitative metrics (ROUGE-L and BERTScore).

Open-source models were selected because the study data included sensitive real-world clinical conversations, and ensuring full deidentification of personally identifiable information would have introduced additional privacy risks and administrative burdens.

<u>Fourth step: Performance assessment</u>

Each AI-generated summaries from 72 transcribed patient-clinician consultations using zero-shot and few-shot prompts (1–3 examples). The outputs were evaluated against the gold-standard PCS using quantitative metrics such as ROUGE-L[14] for lexical similarity and BERTScore[15] for semantic alignment. For all models, hyperparameters were standardized, with temperature fixed at 1 and the maximum number of new tokens set to 2048, as the gold-standard PCS did not exceed this length.

ROUGE-L was calculated using the standard F1 formulation, which measures the longest common subsequence between the model output and the reference summary; scores range from 0 to 1, with higher values indicating greater overlap (e.g., a ROUGE-L of 0.2 reflects relatively limited lexical similarity). BERTScore, also normalized between 0 and 1, was computed using contextual embeddings to capture semantic similarity; higher scores reflect closer alignment in meaning between system and reference summaries.

In the descriptive analysis, we first quantified the amount of patient-centered information present in the source transcripts. Each consultation was manually labeled according to our patient-centered framework, and the number of patient-centered statements per encounter was recorded. Across the 72 consultations, this number ranged from 0 to 16.

A qualitative assessment was performed using the framework proposed by Van Veen et al[16], which we adapted to evaluate summaries across five domains: correctness, completeness, conciseness, patient-centeredness, and fluency. To assess patient-centeredness, we identified which elements of the PCS framework were present in each summary using the labeled transcripts and quantified the amount of patient-centered content included. Five pilot summaries were then rated and discussed to calibrate the approach, and afterward two reviewers (MLJ and AGC) independently rated all summaries. This assessment was conducted in a blinded strategy, as evaluators did not know whether the summaries had been generated by the model or by human experts. For this purpose, summaries were anonymized, randomized, and labeled only as Summary A and Summary B. Each domain was scored on a scale from -5 to +5, where negative values favored the gold-standard PCS, positive values favored the AI-generated summaries, and 0 indicated no difference; directionality was assigned after unblinding to identify which summary corresponded to each source. Scoring was guided by predefined rules: for correctness, we reviewed whether the information in each summary was factually accurate according to the transcript. After unblinding, instances where the model introduced information that was not present in the source, such as inventing medications or clinical details without supporting context, were classified as hallucination and rated -5 if more than two instances were present or -3 if one instance was identified. For patient-centeredness, we compared summaries with tagged transcripts and assigned 0 if both contained a similar amount of patient-centered content, -3 or +3 if one included less than half of the tagged content, and -5 or +5 if one included more than half, depending on which summary achieved this.

To assess interrater reliability between the two evaluators (MLJ and AGC), we conducted a pilot evaluation using three videos, each rated across five parameters on an ordinal scale ranging from -5 to +5. Interrater agreement was quantified using the quadratic weighted Cohen's kappa coefficient, which accounts for the ordinal nature of the data by assigning smaller penalties to minor rating differences. The resulting kappa value was 0.84, indicating excellent agreement[17].

**Ethical Considerations**

The study was conducted in accordance with the International Conference on Harmonisation Good Clinical Practice guidelines. The study protocol was approved by the Institutional Review Board (IRB#25-007747) and included approval for the secondary analysis of data derived from the atrial fibrillation clinical trial.

Written informed consent was obtained from all participants involved in both the PPIE groups and the original clinical trial.

**RESULTS**

**Patient-centered summarization framework**

Through the PPIE participants, including eight clinicians and ten patients representing diverse age groups (**Table 1**). We identified essential themes to guide patient-centered summarization according to patients (**Table 2 and Supplementary Material 3**) Their insights informed the development of a structured framework with five domains: (1) lifestyle and daily routines, (2) support systems and beliefs, (3) health goals and challenges, (4) emotional and mental health, and (5) sources of meaning.

**Table 1. Demographics of PPIE Participants**

| Group | Age Range | Gender (% of group) | Total (% of all) |
|---|---|---|---|
| General Practitioners | N/A | N/A | 7 (38.9%) |
| District Nurses | N/A | N/A | 1 (5.5%) |
| Young Patients | 18–30 | 2 Male (66.7%), 1 Female (33.4%) | 3 (16.7%) |
| Middle-aged Patients | 31–60 | 1 Male (33.3%), 2 Female (66.7%) | 3 (16.7%) |
| Older Patients | 61–91 | 3 Male (75%), 1 Female (25%) | 4 (22%) |

**Table 2 - Evaluation of Large Language Models Using the Patient-Centered Framework. Themes identified from the patient and public involvement (PPIE) interviews, differentiated by older, middle-aged, and younger patients. The table summarizes perspectives across five domains: lifestyle and daily routines, support systems and access, events and life stressors, care preferences, and sources of meaning/value.**

| Category | Older Patients (61–91 years) | Middle-aged Patients (31–60 years) | Younger Patients (18–30 years) |
|---|---|---|---|
| **Lifestyle & Daily Routines** | Active in managing health (e.g., monitoring BP, volunteering) [Patient A, C] | Shared limited lifestyle details (e.g., physical job, retired builder) [Patients H, I] | Limited lifestyle information; concerns about chronic pain [Patient E] |
| **Support Systems & Access** | Reliance on friends for transport; more care needed from loved ones [Patient C] | Family support during crises; challenges with relatives' health [Patients H, I] | Emotional support from family/friends; support sometimes is hard because close friends are often busy [Patient D] |
| **Events & Life Stressors** | Bereavement and chronic illness impact | Bereavement and family illness noted but not shared often | Extrinsic factors like trauma, relationships, work stress |
| **Care Preferences** | Preference for continuity and face-to-face care; avoid technology for engagement [Patient J] | Limited specific preferences | Specific preferences based on condition (e.g., medication avoidance) [Patients D] |
| **Sources of Meaning/Value** | Community contributions, self-management | Less emphasis on sources of meaning unless stressed | Importance of work and family as motivators |

**1. Lifestyle and Daily Routines**

Older patients emphasized the importance of documenting daily routines as indicators of independence and health engagement. For example, they shared efforts to manage blood pressure, engage in volunteering, or maintain physical activity despite mobility challenges. In contrast, middle-aged and younger patients provided fewer details unless health issues clearly affected their routines. When symptoms disrupt daily functioning, such as chronic pain or fatigue, younger patients consider lifestyle information more relevant to record.

**2. Support Systems and Beliefs**

Across age groups, patients valued recording the presence and quality of social support, especially during health crises. Older patients described logistical challenges such as transportation and reliance on friends, while younger and middle-aged patients referred to emotional support from family as relevant though sometimes inadequate. These narratives underscored how both practical and emotional support systems influenced access to care and coping strategies.

**3. Health Goals and Challenges**

Patients of all ages described life stressors affecting their physical and emotional well-being. Younger participants highlighted the impact of trauma, relationship conflict, caregiving roles, and system-level stressors (e.g., child protection fears). Older patients emphasized bereavement or caregiving burdens as key stressors. Some participants expressed a desire for clinicians to recognize how these challenges affected motivation, resilience, and engagement with care.

### 4. Emotional and Mental Health

Preferences for emotional support and mental health disclosure varied. Some younger patients shared significant psychological distress, including suicidality or depression linked to life events. Others wanted clinicians to recognize emotional triggers but preferred sensitive handling. Older patients voiced frustration when distress was dismissed, especially when their usual coping mechanisms failed. The findings revealed that acknowledging emotional states is crucial to understanding health behavior and care needs.

### 5. Sources of Meaning and Personal Values

Older patients identify meaning through community engagement, independence, and contribution, with some explicitly linking these to care preferences, such as avoiding residential care or choosing not to be resuscitated. Middle-aged and younger participants are less often named sources of meaning, unless those values (e.g., work, parenting) were under threat. When identified, these sources of meaning shaped their outlook, motivation, and desire for personalized care planning.

Clinicians also expressed strong support for integrating person-centered information into clinical summaries, emphasizing the importance of understanding patients' functional baselines, psychosocial context, and emotional wellbeing. Many described their role not only as medical providers but also as informal therapists or coaches, often encountering complex cases rooted in bereavement, loneliness, trauma, or caregiving stress. Several clinicians advocated for routinely documenting elements such as support networks, mental health history and care preferences (e.g., avoiding invasive treatments), as well as ICE (patients' Ideas, Concerns, and Expectations).

At the same time, clinicians noted practical considerations for summarization and implementation. Most preferred brief, structured summaries using bullet points, incorporating psychosocial data when relevant. Suggestions for system integration included embedding person-centered content into existing EHR tabs or flagging key insights with visual icons. Some clinicians proposed advanced uses of AI to visualize longitudinal patterns of psychosocial and health-related events or to generate personalized follow-up letters that were humane and accessible. However, a few raised philosophical concerns about whether AI could meaningfully capture the relational and dynamic aspects of healing, particularly the transformative role of the patient-clinician interaction itself.

<u>Baseline characteristics</u>

A total of 72 transcribed patient-clinician consultations from the atrial fibrillation trial were analyzed, with general characteristics described in **Table 3**. The mean patient age was 70 years (SD 11), and 44.4% were female. Most consultations took place at Mayo Clinic Rochester (34.7%). While all encounters were related to atrial fibrillation, they occurred in different clinical settings, most commonly cardiology appointments (55.6%), followed by thrombophilia clinic (12.5%). The most prevalent comorbidity was a history of hypertension, reported in 87.5% of participants.

**Table 3. Demographic and clinical characteristics of patients participating in the atrial fibrillation conversations. SD: Standard Deviation.**

| General Characteristics | Total (N=72) |
|---|---|
| **Age (y):** Mean (SD) | 70 (11) |
| **Gender**, n (%) | |
| Female | 32 (44.4%) |
| Male | 40 (55.6%) |
| **BMI:** Mean (SD) | 33.0 (7.71) |
| **Location**, n (%) | |
|     Mayo Clinic, Rochester, Minnesota | 25 (34.7%) |
|     Park Nicollet, Minnesota | 19 (26.4%) |
|     Hennepin County Medical Center | 12 (16.7%) |
|     Alabama | 4 (5.6%) |
|     Mississippi | 12 (16.7%) |
| **Appointment Length (min):** Mean (SD) | 33.8 (17.5) |
| **Appointment Type**, n (%) | |
|     Emergency Department | 1 (1.4%) |
|     Primary Care/ Family Medicine | 4 (5.6%) |
|     Inpatient | 4 (5.6%) |
|     Cardiology | 40 (55.6%) |
|     Thrombophilia Clinic | 9 (12.5%) |
|     Parkside Ambulatory | 2 (2.8%) |
|     Brooklyn Center Ambulatory center | 7 (9.7%) |
|     Anticoagulation Clinic | 3 (4.2%) |
|     Other | 2 (2.8%) |
| **History of:** | |
|     Hypertension, n (%) | 63 (87.5%) |
|     Congestive Heart Failure, n (%) | 13 (18.1%) |
|     Stroke, n (%) | 12 (16.7%) |
|     Stroke Type, n (%) | |
|       TIA | 3 (25%) |
|       Ischemic stroke | 9 (75%) |
|     Vascular Disease, n (%) | 21 (29.2%) |
|     Diabetes, n (%) | 30 (41.7%) |
|     Renal Disease, n (%) | 10 (13.9%) |
|     Liver Disease, n (%) | 4 (5.6%) |
|     MI, n (%) | 5 (6.9%) |
|     PAD, n (%) | 2 (2.8%) |

Quantitative evaluation

In the quantitative evaluation, the best performance in the zero-shot setting was achieved by Qwen3-8B on ROUGE-L (0.189), and Llama-3.1-8B on BERTScore (0.673). As few-shot attempts were introduced, performance improved across all models. With one example, Llama-3.1-8B outperformed all others, achieving a ROUGE-L of 0.201 and a BERTScore of 0.680. The best-performing model overall was Llama-3.1-8B with three-shot prompting, which achieved a ROUGE-L of 0.206 and a BERTScore of 0.683.

Qualitative evaluation

Only one consultation had no patient-centered content, while among the remaining 71, 10 included a single domain and 61 incorporated multiple domains. Patient-centered topics spanned lifestyle and daily routines, support systems, life stressors, care preferences, and sources of meaning.

The qualitative human evaluation compared the AI-generated summaries from the best-performing model against the gold-standard PCS (**Figure 2**). The two approaches performed similarly in terms of completeness (-0.6 ± 2.1) and fluency (-0.6 ± 2.1). However, the gold-standard PCS were rated significantly higher for correctness (-2.5 ± 1.9), as the model occasionally hallucinated procedures and medications not mentioned in the consultations. Conciseness slightly favored the model (0.6 ± 2.2). The most substantial difference was in patient-centeredness, where gold-standard PCS vastly outperformed the model (-4.0 ± 1.5), reflecting their superior ability to capture the nuanced details patients shared about their daily routines, support networks, and personal values. On average, transcripts contained (3.6 ± 2.5) patient-centered domains, whereas AI-generated summaries captured only (0.8 ± 0.9), and nearly half (35/72, 49%) contained none.

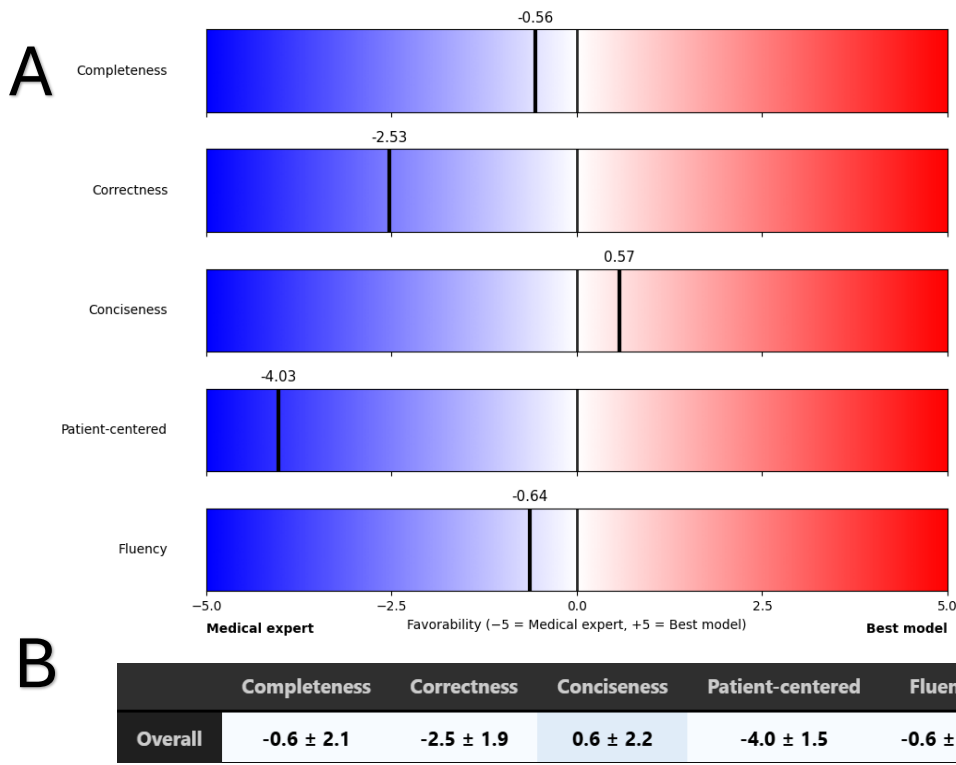

**Figure 2. Summary of qualitative evaluation of the AI-generated summaries with Llama-3.1-8B (A)** Diverging bar plots showing mean favorability ratings for each evaluation dimension. Ratings range from -5 (indicating strong preference for the *medical expert*) to +5 (indicating strong preference for the *Best model*). Vertical black lines mark the average score for each dimension, with more negative values favoring the medical expert and more positive values favoring the model. **(B)** Summary table reporting the overall mean ± standard deviation (SD) for all five evaluation dimensions: *Completeness, Correctness, Conciseness, Patient-centered,* and *Fluency*. Negative values indicate preference for the medical expert, while positive values indicate preference for the best model.

### DISCUSSION

This study shows the potential and value of using PPIE methodology for developing patient-centered frameworks for AI clinical summarization as it streamlines its development while ensuring it truly represents what "patient-centered" means for both patients and clinicians while ensuring clinical utility. Patients highlighted the importance of including lifestyle routines and daily functioning, social support systems, recent life stressors, and personal values or care preferences whereas clinicians prioritized the need to capture patient's functional baseline and psychosocial context in a structured, clinically accessible format, such as bullet points. Using this standard as PCS, quantitative and qualitative evaluation showed that even the best-performing LLMs demonstrated poor alignment with the gold-standard PCS. The domains identified in our PCS framework align with prior frameworks aiming to incorporate holistic patient perspectives into care documentation and planning[18,19]. In our framework, older participants emphasized independence and daily functioning as central concerns, reflecting well-documented age-specific differences in stress perception and priorities. As highlighted by the American Psychological Association, declines in functional abilities increase reliance on environmental supports to maintain autonomy and quality of life, reinforcing older adults' focus on daily routines and health-related function[20]. In contrast, younger participants in our study discussed emotional burdens and system-level stressors, including work, finances, and relationships, patterns consistent with evidence that younger adults report higher total daily stress and are more affected by interpersonal and occupational demands[21,22].

Clinicians also expressed a desire to systematically capture person-centered data. They viewed understanding patient's functional baselines, emotional challenges, and preferences as central to care, according to the biopsychosocial model widely encouraged in primary care literature[23]. However, clinicians also raised the need to generate summaries with brief, structured documentation formats and suggested integrating person-centered content into discrete sections of the EHR (e.g., icons or lifestyle tabs). Previous studies have also mentioned that clinicians have expressed the need for EHR systems to facilitate intuitive, less burdensome documentation interfaces and to allow for both structured and narrative content, supporting both efficient data entry and the nuanced, person-centered aspects of care[24].

In our evaluation, gold-standard PCS were notably stronger in patient-centeredness and correctness compared to AI-generated summaries, underscoring key limitations of current LLMs in clinical summarization. Our evaluation of five open-source LLMs revealed poor performance in generating PCS compared to the gold standard. Despite modest gains with few-shot prompting, the models showed limited semantic alignment (BERTScore) with gold-standard PCS. This reflects a major limitation in current LLM development: most are trained in general-purpose data and are not familiar with the type of personal, emotional, and value-based content essential for patient-centered care[25]. This limitation is also related to the nature of the training data, which prioritizes pathophysiological detail over patients' preferences or psychosocial context and reproduces this imbalance, generating summaries focused on biomedical issues and the clinician's perspective, with limited attention to patients' values and concerns. The lack of dedicated datasets, annotation standards, and evaluation tools for PCS further contributes to this problem, making it difficult for models to capture the social, emotional, and functional information that matters most to patients and clinicians.

Several limitations must be considered. The study design, which involved developing the framework with a UK population and testing it on US consultation data, was a pragmatic decision driven by the availability of a high-quality, IRB-approved dataset. While the core principles of patient-centeredness are universal, the specific expression and prioritization of patient concerns can be shaped by cultural and systemic contexts. For instance, the priorities of a patient within the UK's National Health Service may differ from those of a patient navigating the US insurance system, where financial considerations can be a more prominent part of the illness experience[26]. Therefore, while our UK-derived framework serves as a robust foundational model, its application to other healthcare systems might require minor adjustments for optimal use. Future research should focus on validating and adapting this framework across diverse settings.

Other limitations include the exclusion of larger proprietary models and the use of fine-tuning. The quantitative metrics used also have known weaknesses; ROUGE and BERTScore primarily measure lexical and semantic similarity and can fail to capture differences between semantic alignment of clinical issues and of patient-centered issues, which reinforces the importance of our qualitative human evaluation. In addition, our findings are based on consultations for a single clinical condition and exclude more structured encounters, which may limit generalizability. Finally, our gold-standard PCS were generated by combining patient-centered themes identified through PPIE with clinically meaningful information, providing a practical and reproducible benchmark. Although this approach reduced variability through consensus among annotators, an ideal standard would involve developing the summary immediately after each clinical encounter and obtaining direct patient feedback to verify whether the PCS accurately reflect their priorities and experiences.

## CONCLUSION

This study shows that producing truly patient-centered clinical summaries remains a major challenge for current open-source LLMs. Using a mixed-methods framework grounded in patient and clinician input, we identified the key domains of a PCS and demonstrated that a PPIE-driven approach can streamline development while ensuring summaries reflect both perspectives and remain clinically relevant. The tested models failed to capture these dimensions, particularly factual correctness and nuanced patient values, with ROUGE-L and BERTScore results well below human benchmarks. Our work provides a validated framework, a reference standard for future research, and a pathway toward patient-centered AI systems that better support shared decision-making.

## ACKNOWLEDGEMENTS

We would like to acknowledge the patients and general practitioners who participated in the Patient and Public Involvement and Engagement groups in the United Kingdom, whose insights were essential to developing the patient-centered framework.

## DATA AVAILABILITY

The data generated in this study include identifiable patient–clinician encounter recordings and linked electronic health record data. Due to privacy and ethical restrictions, this data cannot be made publicly available. De-identified data may be made available upon reasonable request to the corresponding author and with appropriate institutional approvals.

## AUTHORS' CONTRIBUTIONS

MLJ, AGC, KG, JPB, and OJPP conceived the study idea. MLJ and AGC led the protocol drafting, system design, and manuscript preparation. KG, SA, SL, and ML led the Patient and Public Involvement (PPI) groups in the United Kingdom. FL, CW, KGM, LVA, CPV, SSSB, and SB contributed to the development of the annotation guideline and creation of the gold-standard PCS. DTT, OJPP, and MAZ worked on prompting design and data analysis. MEB performed the statistical analyses. JPB and OJPP provided overall supervision, clinical oversight, and critical revision of the manuscript. All authors reviewed and approved the final version of the manuscript.

## CONFLICTS OF INTEREST

DTT is a consultant for Immunovant but reports no conflicts of interest related to this publication. All other authors declare that they have no conflicts of interest related to this work

## FUNDING

This work was supported by the National Institute for Health and Care Research (NIHR; Award ID: NIHR207380) and the Torbay Medical Research Fund (Grant 145).

**ABBREVIATIONS**

- AI: Artificial Intelligence
- BERTScore: Bidirectional Encoder Representations from Transformers Score
- CARE: Consultation and Relational Empathy
- EHR: Electronic Health Record
- F1: F-measure (harmonic mean of precision and recall)
- IRB: Institutional Review Board
- LLM: Large Language Model
- PCS: Patient-Centered Summaries
- PPIE: Patient and Public Involvement and Engagement
- ROUGE-L: Recall-Oriented Understudy for Gisting Evaluation–Longest Common Subsequence
- SD: Standard Deviation

**Supplementary material S1. Annotation Guideline**

Instructions

Step #1:

- Elaborate the patient-centered (PC) summaries first. [Read about patient-centered information – see word documents]
- Generate summary by **only** watching the video – do not use/read the transcripts while the video is playing. You can read the transcripts, but you should stop the video when you do so.
- Do not forget to highlight in yellow the PC information on the PC summaries. This will help you in the second step and it is there to help all of us (the researchers) where the PC information is.
- Outline of the PC summary

**Background [bg1]:** Past medical history, including an ongoing medical condition such as Atrial fibrillation, and age. For instance, patient goes to a follow-up appointment of his AFib medication; this will be considered both as background and issues. Do this in chronological order based on how it is mentioned on the video.

> **EVERY medical condition** that has any type of information (e.g., medication, PC information, follow-up info, doctor who is following the condition) should go in one bullet.
>
> **All medical conditions** that are just listed without any additional information should go all of them in ONE bullet point.
>
> **EVERY Medication** that is related to a medical condition should be in the bullet that corresponds to the medication.
>
> **All medications** that are just listed and not related to medical conditions spoken during the consultation should be all lumped together (grouped into one bullet).
>
> **You can have BG with and without PC information.**
> - Background with PC information
> - Background without PC information

**Issues [issue1]:** In chronological order.

- Issues can have PC information.
- Issues without PC information.

**Plan [plan1]:** Write the plan based on the orders of the issues.

**Step #2:**

PLEASE BE AS PRECISE AS POSSIBLE WHEN TAGGING YOUR SENTENCES. Do not include unnecessary information.

- Tag the parts of the transcripts that helped you elaborate the PC summary.
- You will do this by reading your patient-centred summary and identifying the sentences from the transcript the helped you elaborate this summary.

- The minimum unit for tagging is a sentence. Sentences are defined as those that end with a period "." or exclamation marks ("?", "!").

**A tag can never end within a sentence. It should end at the end of the sentence (read the definition of sentence on the previous point, please)**

- You need to tag these sentences based on the following nomenclature.
    - Background -> Tags -> no order
        - beginning -> [bg1]
        - Final -> [/bg1]
    - Issues - order
        - Organise issues by number. Tags
- [issue1]
- [/issue1]
    - Plan - order
        - [plan1]
        - [/plan1]

**Supplementary material S2. Prompt for Patient-Centered Clinical Summaries**

You are an experienced doctor who generates patient-centered clinical summaries of patient-doctor conversations. You must structure the clinical summaries into three sections: Background, Issues, and Plans. Use the following format:

**1. Background**

Summarize the patient's historical medical context in concise bullet points and Group related details logically. Include:

- **Past Medical History**: Diagnosed conditions, surgeries, or chronic illnesses.

- **Medications**: Current prescriptions, dosages, and duration.

- **Social History**: Relevant lifestyle factors (e.g., smoking, alcohol use, occupation).

- **Allergies**: Documented drug or environmental allergies.

**2. Issues**

List the primary medical concerns addressed during the conversation in bullet points. Group related details logically to minimize the number of bullet points. Each bullet point must provide a thorough description of the medical concern.

You must be more descriptive in this section with more narrative content.

**3. Plan**

List diagnostic or treatment plans that patients or doctors must do after consultation. Group related details logically to minimize the number of bullet points.

**Supplementary material S3**

Illustrative examples of how patients in different age groups described their experiences across thematic categories (Lifestyle & Daily Routines, Support Systems & Access, Events & Life Stressors, Care Preferences, and Sources of Meaning/Value), corresponding to the summary presented in Table 2.

**Lifestyle & Daily Routines:**
- *Patient A* (older) monitored blood pressure and supported pharmacy routines.
- *Patient C* (older/middle-aged) emphasized healthy lifestyle and volunteering.
- *Patient H* (middle-aged) described work as physically demanding; *Patient I* (middle-aged) retired from building work.
- *Patients F and G* (younger) did not share lifestyle details.
- *Patient E* (younger) reported chronic pain affecting daily life.

**Support Systems & Access:**
- *Patient J* (older) Issues accessing General Practitioners.
- *Patient C* (older) More care needed for/from loved ones.
- *Patient H* (middle-aged) mentioned mother's health issues; *Patient I* (middle-aged) reported dealing with family health problems.
- Patient D (younger) described his close friends as often being busy; Patient E (younger) noted that his wife provided limited support, not the needed kind; Patient G (younger) identified his mother as a key source of support.

**Events & Life Stressors:**
- *Patient B* (older) had suicidal risk due to mobility.
- *Patient H* (middle-aged) affected by mother's Chronic Obstructive Pulmonary Disease; *Patient I* by aunt's dementia.
- *Patient F* (younger) experienced past traumas and relationship difficulties, *Patient D* reported fear of social services.

**Care Preferences:**
- *Patient J* (older) preferred face-to-face care and avoided technology for engagement.
- *Patients C, F, G, H, I* (middle-aged) expressed no specific preferences.
- Patient D (younger) avoided Selective Serotonin Reuptake Inhibitors and required encouragement; Patient E emphasized the need for clear medical history summaries.

**Sources of Meaning/Value:**
- Patient A (older) found meaning in the community, Patient C in mental health.
- *Patient H* (middle-aged) emphasized community connectedness.
- Patient D (younger) described children as motivators; Patient E cited a small business as a source of distraction and meaning.